# PLOS ONE

RESEARCH ARTICLE

# Directions in abusive language training data, a systematic review: Garbage in, garbage out

Bertie Vidgen[1]*, Leon Derczynski[2]

**1** The Alan Turing Institute, London, United Kingdom, **2** Department of Computer Science, IT University of Copenhagen, Copenhagen, Denmark

* bvidgen@turing.ac.uk

## Abstract

Data-driven and machine learning based approaches for detecting, categorising and measuring abusive content such as hate speech and harassment have gained traction due to their scalability, robustness and increasingly high performance. Making effective detection systems for abusive content relies on having the right training datasets, reflecting a widely accepted mantra in computer science: Garbage In, Garbage Out. However, creating training datasets which are large, varied, theoretically-informed and that minimize biases is difficult, laborious and requires deep expertise. This paper systematically reviews 63 publicly available training datasets which have been created to train abusive language classifiers. It also reports on creation of a dedicated website for cataloguing abusive language data hatespeechdata.com. We discuss the challenges and opportunities of open science in this field, and argue that although more dataset sharing would bring many benefits it also poses social and ethical risks which need careful consideration. Finally, we provide evidence-based recommendations for practitioners creating new abusive content training datasets.

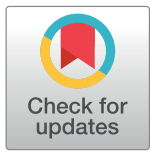

**Data Availability Statement:** The data are detailed in the relevant table in the appendix.

**Funding:** This study was supported by Wave 1 of The UKRI Strategic Priorities Fund under the EPSRC Grant EP/T001569/1 and Grant TPS2019 100065 awarded to BV, as well as an internal grant from the IT University of Copenhagen on Abusive Language Detection awarded to LD.

**Competing interests:** The authors have declared that no competing interests exist.

## Introduction

Abusive online content, such as hate speech and harassment, has received substantial attention from academics, policymakers and big tech companies. Left unchallenged, abusive content risks harming those who are targeted, toxifying public discourse, exacerbating social tensions and threatening exclusion of targeted groups from public spaces. Systems which can accurately detect and classify online abuse at scale, in real-time and without bias would substantially help to address these issues.

Making effective detection systems for abusive content relies on having the right training datasets. This follows conventional wisdom in data-intensive research: Garbage In, Garbage Out. However, creating training datasets that have the size, variation, grounding and low-bias required is a serious undertaking: it requires volumes of painstaking labour and specific, deep expertise. We examine this issue by conducting a systematic review of publicly available datasets for abusive content detection.

The first section surveys previous reviews and present the four research aims which guide this paper. The second section summarizes the literature review methodology (using

PRISMA). Then, the third section gives an in-depth analysis of 63 publicly-available training datasets for abusive content detection, exploring how tasks have been described and what taxonomies have been constructed, what the datasets contain and how they were annotated. In the fourth section, we discuss the challenges and opportunities of open science and more dataset sharing in this field, and elaborate different ways of sharing training datasets, including the new website hatespeechdata.com. In the final section, we draw on our findings to provide evidence-based recommendations for practitioners using training datasets to create new abusive content detection systems.

## Background

The amount of research examining the social and computational aspects of abusive content detection has expanded prodigiously in the past five years. This has been driven by growing awareness of the importance of Internet-based communications within society [1], greater recognition of the harms caused by online abuse [2], and policy and regulatory developments, such as the EU's Code of Conduct on Hate, the UK Government's 'Online Harms' white paper [3], Germany's NetzDG laws, the Public Pledge on Self-Discipline for the Chinese Internet Industry, and France's anti-hate regulation [2]. In 2020 alone, at least three computer science venues will host workshops on online hate (TRAC and STOC at LREC, and WOAH at EMNLP), and a shared task at 2019's SemEval on online abuse detection reported that 800 teams downloaded the training data and 115 submitted detection systems [4]. At the same time, social scientific interventions have also increased, deepening our understanding of where online abuse manifests, how it spreads spreads, and how its harmful impact can be mitigated and challenged [5–7].

Having a robust way of detecting and measuring online abuse is important for at least three downstream tasks: (1) analysing its prevalence dynamics, patterns and causes within social scientific research, (2) moderating content that might contravene a platforms' Terms of Service and (3) identifying vulnerable users and victims in need of targeted support and interventions. This means having a method which can handle the volume of content produced, shared and engaged with online. Traditional qualitative methods cannot scale to handle the hundreds of millions of posts which appear on each major social media platform every day, and introduce various inconsistencies and biases [8]. That said, human-based moderation is still an integral part of most platforms' approaches to tackling online abuse; a 2020 report showed that Facebook had over 15,000 content moderators (many of whom work for outsourcing companies) with the author alleging that there is an immediate need for 30,000 [9]. However, having large numbers of moderators risks their health and well-being and can be economically inefficient [10, 11]. As a result, many companies have combined both approaches, embedding data-intensive technologies in hybrid systems where they are often used to filter and surface content for human moderators.

The use of AI and related technologies to moderate undesirable online content, such as abuse, has generated considerable debate, particularly in light of discussions about algorithmic bias, fairness and predictive decision making [12–14]. Tarleton Gillespie argues that part of the challenge is that AI is seen as an end-in-itself—instead, he argues that AI should be used 'not as a replacement for repetitive human judgments but to enhance the intelligence of the moderators, and the user communities, that must undertake this task.' [15] [p. 5]. This reflects a wider recognition that, whilst algorithmic content moderation systems are problematic, they are also an unavoidable part of the solution. For instance, Gorwa et al. write that, 'algorithmic moderation has become necessary to manage growing public expectations for increased platform responsibility, safety and security on the global stage' [16] [p. 1]. In this regard, the key task for researchers, companies and governments is to ensure that detection systems are created which are fair and accurate, and used in a responsible way, rather than to try stop their use altogether.

Considerable advances have been made in how computational tools are used to classify and detect online abuse, drawing on research in machine learning, Natural Language Processing (NLP) and statistical modelling. Sophisticated architectures, features and processes have been used, such as contextual word embeddings, paragraph embeddings, graph embeddings and dependency parsing. Despite their many differences [17], nearly all systems for detecting online abuse rely on a *training dataset*, which is used to teach the system what is and is not considered abuse. This is a crucial aspect of the machine learning process, and a key point at which social scientific insights are needed [18]. However, despite its importance, there is a lacuna of research on this part of abusive content detection. Indeed, although several general reviews of the field have been conducted, none have addressed the role of training datasets in sufficient breadth or depth or specifically reviewed their role. This is surprising given (i) their fundamental importance in creating detection systems and (ii) growing awareness that several existing datasets suffer from many flaws [19–21]. Relevant previous work includes:

- Schmidt and Wiegand conduct a comprehensive review of research into the detection and classification of abusive online content. They discuss training datasets, stating that 'to perform experiments on hate speech detection, access to labelled corpora is essential' ([17], p. 7), and briefly discuss the sources and size of the most prominent existing training datasets, as well as how datasets are sampled and annotated. Two key challenges with existing datasets are identified. First, 'data sparsity': many training datasets are small and lack linguistic variety. Second, metadata (such as how data was sampled) is often not adequately reported even though it is crucial for understanding dataset biases ([17], p. 6).

- Waseem et al. [22] outline a typology of detection tasks, based on a two-by-two matrix of (i) identity- versus person- directed abuse and (ii) explicit versus implicit abuse. They emphasise the importance of high-quality datasets, particularly for more nuanced expressions of abuse: 'Without high quality labelled data to learn these representations, it may be difficult for researchers to come up with models of syntactic structure that can help to identify implicit abuse.' ([22], p. 81)

- Jurgens et al. [23] conduct a critical review of hate speech detection research, and note that 'labelled ground truth data for building and evaluating classifiers is hard to obtain because platforms typically do not share moderated content due to privacy, ethical and public relations concerns.' ([23], p. 3661) They argue that the field needs to 'address the data scarcity faced by abuse detection research' in order to better address more complex research issues and pressing social challenges, such as 'develop[ing] proactive technologies that counter or inhibit abuse before it harms' ([23], pp. 3658, 3661).

- Vidgen et al. review challenges and frontiers in abusive content research, and describe several limitations with existing training datasets, most noticeably how 'they contain systematic biases towards certain types and targets of abuse.' [11] [p.2]. They describe three issues with the quality of datasets: degradation (whereby datasets decline in quality over time), annotation (whereby annotators often have low agreement, indicating considerable uncertainty in class assignments) and variety (whereby 'The quality, size and class balance of datasets varies considerably.' [p. 6]).

- Waqas et al. conduct a scientometric analysis of research into online hate [24]. They analyse papers from Web of Science until March 2019, and show a sharp increase in the volume of research, from 50 publications in 2005 to 550 in 2018. The top papers, institutions and several distinct clusters of research are shown, providing a broad lens on the development of the field. However, Waqas et al. also show that, despite the cross-disciplinary nature of research into online hate, most publications originate from one disciplinary area: psychology and psychiatry. Given this, they argue that there needs to be a shift towards more practical considerations, such as 'design[ing] both technological and non-technological solutions to identify and curb online hate.' [24] [p.17]

- Chetty and Alathur review the use of Internet-based technologies and online social networks to study the spread of hateful, offensive and extremist content [25]. Their review covers both computational and legal/social scientific aspects of hate speech detection, and outlines the importance of distinguishing between different types of group-directed prejudice. However, they do not consider training datasets in any depth.

- Fortuna and Nunes [26] provide an end-to-end review of hate speech research, including the motivations for studying online hate, definitional challenges, dataset creation/sharing, and technical advances, both in terms of feature selection and algorithmic architecture ([26], 2018). They delineate between different types of online abuse, including hate, cyberbullying, discrimination and flaming, and add much needed clarity to the field. They show that (1) dataset size varies considerably but they are generally small (mostly containing fewer than 10,000 entries), (2) Twitter is the most widely-studied platform, and (3) most papers research hate speech *per se* (i.e. without specifying a target). Of those which do specify a target, racism and sexism are the most researched. However, their review focuses on publications rather than datasets. This is problematic because the same dataset might be used in multiple studies, limiting the relevance of their review for understanding just training datasets.

- Davidson et al. and Sap et al. draw attention to the importance of training datasets, showing that systems which are trained on some of the most widely-used datasets have performance biases against content written in African-American English versus standard English vernaculars [19, 21]. Sap et al. argue that 'Our findings corroborate the existence of racial bias in the toxic language datasets and confirm that models propagate this bias when trained on them' [21] [p. 1671]. These data-driven analyses emphasize the importance of datasets and the need to critically interrogate this part of the machine learning process.

- Several classification papers also discuss the most widely used datasets, including Davidson et al. [27] who describe five datasets, and Salminen et al. who review 17 datasets and describe four in detail [28].

## Objectives

This paper addresses the lack of existing research into the quality, coverage and focus of training datasets for online abuse detection, providing a systematic review of the available sources. This is a crucial issue as the field of online abusive content detection matures and tackles more complex research challenges, such as multi-platform, multi-lingual and multi-target abuse detection. It also has real world implications as detection systems are increasingly deployed in 'the wild' for social scientific analyses and content moderation [2]. As such, this systematic review paper has four research aims.

1. *Research Aim One*: to provide an in-depth and critical analysis of the available training datasets for abusive online content detection.

2. *Research Aim Two*: to map and discuss ways of addressing the lack of dataset sharing, and as such the lack of 'open science', in the field of online abuse research.

3. *Research Aim Three*: to introduce the website hatespeechdata.com, as a way of enabling more dataset sharing.

4. *Research Aim Four*: to identify best practices for creating abusive content training datasets.

## Method for literature survey

This study's selection criteria, protocol, and methodology follow the PRISMA framework [29, 30], based on the guidance provided by [31]. The process we followed is shown in S1 Fig.

### Eligibility criteria

The criteria for including datasets in this study is whether they contain annotations for *online abuse*, and as such can be used to train detection/categorization systems. We construe 'abuse' to include hate speech, interpersonal harassment, toxic language, trolling, aggression, bullying and profanity [11, 32]. Primarily, we focus on content that makes attacks against identities/ groups (such as attacks against women, ethnic minorities etc.), as well as personal attacks and highly aggressive forms of language. We construe 'online' in a relatively broad way, including any digital medium, such as social media platforms, digital forums, Internet-enabled mobile phone communications, and comments sections on news sites. We did not place any restrictions in terms of language or year for identifying relevant datasets. Our only publication venue requirement was that datasets were described in a publication accepted by a recognised conference or journal. We only include datasets that are publicly accessible without restrictive license, either directly (i.e. content and annotations are provided publicly through a repository, such as GitHub or Zenodo) or on request (i.e. once a Terms of Service form is completed, researchers are sent the data). This is because datasets that cannot be downloaded by other researchers cannot be fully scrutinized and do not further data-driven machine learning research, which require first-hand complete access to training sets. Further details of the data surveyed are given in Appx.

### Sources of content

Abusive content training datasets were identified from three sources, providing coverage of a range of academic venues.

1. *Scopus* A cross-disciplinary database of academic research which regularly indexes different sources, primarily journal papers and conference proceedings. Scopus is comparable to Web of Science and Google Scholar in its coverage and, as with Web of Science, has quality control processes that limit inclusion of irrelevant entries [33, 34]. It has been used in previous surveys of abusive content research [26].

2. *ACL Anthology* A database of research in the fields of computational linguistics and natural language processing published by the Association for Computational Linguistics [35, 36]. It includes the proceedings of the first, second and third workshops on abusive language online.

3. *arXiv* An open-access repository of preprints, which are approved before posting but do not undergo full peer review. Publishing early drafts of papers, pre-accepted research and fully accepted research (after its official publication) is a common practice in many natural scientific fields and maths [37, 38].

## Search strategy

We Searched the three databases using a closely related set of keywords which operationalize our eligibility criteria. We explored using other terms, including 'Dangerous', 'harm', 'Discrimination', 'Attack', 'Aggression', 'Stalking', 'Violence', 'Extremist', 'Malicious' and 'Spam' but did not include them as they returned a large number of irrelevant publications. For all three databases we used the same basic search string for abuse, with two changes for Scopus. First, we used wildcards (as this functionality is supported by the platform) whereas for the other two we had to manually write out all plausible variations. Second, because Scopus indexes a far larger range of content we used an additional qualifying clause, limiting the results only to publications which concern online activity. Searches were first implemented in February 2019 and last implemented in February 2020.

1. *Scopus* ("hate*" OR "bully" OR "offensiv*" OR "troll" OR "prejudi*" OR "harass*" OR "personal attack*") AND ("online" OR "digital*" OR "cyber*" OR "social media" OR "internet" OR "web" OR "social network*")

2. *ACL Anthology* ("hate" OR "hates" OR "hateful" or "hating" OR "offensive" OR "offence" OR "offensiveness" OR "troll" OR "trolls" OR "trolling" OR "prejudice" OR "prejudicial" OR "harass" OR "harassing" OR "harassment" OR "bully" OR "bullies" OR "bullying" OR "personal attack*")

3. *arXiv* ("hate" OR "hates" OR "hateful" or "hating" OR "offensive" OR "offence" OR "offensiveness" OR "troll" OR "trolls" OR "trolling" OR "prejudice" OR "prejudicial" OR "harass" OR "harassing" OR "harassment" OR "bully" OR "bullies" OR "bullying" OR "personal attack*")

## Study selection

The publications returned from the three data sources were manually reviewed by two researchers for whether a publicly available abusive content training dataset was described. All publications in which at least one researcher identified a training dataset were selected and then re-reviewed by a third researcher who attempted to collect the dataset. In some cases, datasets were either not actually described in the paper (due to an identification error by the one of the first two researchers) or had not been made publicly available and so were not included. Most publications report on the creation of one abusive content training dataset. However, some describe several new datasets simultaneously or provide one dataset with several distinct subsets of data [39–42]. For consistency, we separate out each subset of data where they are in different languages or the data is collected from different platforms. This is not possible in only one case, which is because the authors combined a small number of monolingual entries from three sources and reported the results in aggregate [43]. We treat this as a single dataset in the study. In total, 63 datasets were identified, all of which were released between 2016 and 2020, as shown in S2 Fig. The list of included datasets is provided in Table 2 in S2 Appendix. Identification took place in the period December 2019 to April 2020.

## Data analysis methodology

The primary goal of our analysis is to understand the content of, and to critically analyse, existing abusive content training datasets [44], rather than to conduct a meta-analysis [44]. As such, many statistical methods and tests commonly used in systematic surveys and analyses, such as an Egger's test, are not appropriate [45]. Furthermore, we do not apply any weighting to the datasets to account for potential publication bias. This is because the total number of datasets is very low and their content is highly heterogeneous. Weighting by any relevant features (e.g. dataset size, age or number of citations) would introduce even more substantial biases. This reflects the particularities of this field, namely that (a) it is relatively new, with the first substantial study reported just in 2012 [46], (b) three datasets dominate the field, especially amongst studies which develop new engineering solutions for abusive content detection [27, 32, 47] and (c) the tradeoff between the size and quality of datasets means that the number of entries is an inappropriate weighting criteria. For these reasons, we do not use statistical methods for publication bias and provide unweighted mean estimates of the relevant characteristics in our analysis.

To provide structure to how we analyse the datasets, we partly adopt the 'data statements' framework put forward by Bender and Friedman [48], as well as other work that provides frameworks, schema and processes for analysing NLP artefacts [49–51]. Data statements are a way of capturing how datasets were created in Natural Language Processing (NLP) research. They formalise how decisions should be made and documented, not only ensuring scientific integrity but also addressing 'the open and urgent question of how we integrate ethical considerations in the everyday practice of our field' ([48], p. 587). In some cases, however, we find that it is not possible to fully recreate the level of detail recorded in an original data statement from how datasets are described in their publications.

## Analysis of training datasets

### Motivations behind datasets

Creating a training dataset for online abuse detection is typically motivated by the desire to address a social problem. These motivations can inform how the taxonomy of abusive language is designed, how data is collected and what instructions are given to annotators. We identify the following motivating reasons, which were explicitly referenced by the creators of at least one of the 63 datasets.

1. *Reducing harm*: Aggressive, derogatory and demeaning online interactions can inflict harm on individuals who are targeted by such content, as well as those who observe it. This has been shown to have detrimental long-term consequences on individuals' well-being, with some vulnerable individuals expressing concerns about leaving their homes following experiences of abuse [52].

2. *Removing illegal content*: Many countries legislate against certain forms of speech, e.g. direct threats of violence. For instance, the EU's Code of Conduct requires that all illegal online hate speech is reviewed within 24 hours, and removed if necessary [53]. Many large social media platforms and tech companies adhere to this code of conduct (including Facebook, Google and Twitter). However, in most cases the abuse that is marked up in training datasets falls short of the requirements of illegal online hate—indeed, as most datasets are taken from public source, most of the illegal content has already been removed.

3. *Improving health of online conversations*: The health of online communities can be severely affected by abusive language. It can fracture communities, exacerbate tensions and even repel users. This is not only bad for the community and for civic discourse in general, it can also negatively impact engagement and thus the revenue of the host platforms. There is a growing impetus to improve user experience and ensure online dialogues are healthy, inclusive and respectful. There is ample scope for improvement: a 2017 study showed that 82% of personal attacks on Wikipedia against editors are not addressed [47].

4. *Reducing the burden on human moderators* Automatic removal of abusive content reduces the amount of harm and strain imposed on both commercial and voluntary content moderators, who are routinely exposed to abusive content, often with insufficient safeguards, and sometimes display symptoms similar to those of PTSD [10]. Automatic tools could help to lessen this exposure, reducing the burden on moderators.

## Detection tasks

Myriad tasks have been addressed in the field of abusive online content detection, reflecting the different disciplines, motivations and assumptions behind research. This has led to considerable variation in what is actually detected under the rubric of 'abusive content', and establishing a degree of order over the diverse categorisations and subcategorisations is both difficult and involves making choices that are necessarily somewhat reductive. Key dimensions which dataset creators have used to frame the detection tasks they tackle include what entity is targeted (e.g. groups vs. individuals), the strength of content (e.g. covert vs. overt), the nature of the abuse (e.g. benevolent vs. hostile sexism), how the abuse manifests (e.g. threats vs. derogatory statements), the tone (e.g. aggressive vs. non-aggressive) and the specific target (e.g. ethnic minorities vs. women). Other important dimensions include the theme used to express abuse (e.g. distinguishing between Islamophobia which relies on tropes about terrorism vs. tropes about sexism) and the use of particular linguistic devices, such as appeals to authority, sincerity and irony. All of these dimensions can be combined in different ways, producing a large number of intersecting tasks.

Due to the variety of social phenomena that they address, existing datasets often cannot be easily combined to form 'mega-datasets'—an otherwise promising avenue for overcoming data sparsity, as in [28]. This problem is exacerbated by the fact that consistency in how tasks are formulated is not the only consideration using datasets interchangeably. From the description of a task, an annotation framework must be developed which converts the conceptualisation of abuse into a set of annotation standards. This formalised representation of the 'abuse' inevitably involves creating shortcuts, imperfect rules and simplifications. If two datasets are developed for the same task but use different annotation frameworks—or, even, use the same annotation framework but apply it differently—then they might still vary considerably and be incommensurable.

Ultimately, how detection tasks in online abuse are defined and operationalized is crucial for how they—and, in turn, systems trained on data for them—are subsequently used to generate new knowledge. For example, while a dataset annotated for hate speech can be used to examine biases in annotation, the reverse is not necessarily true. That is, a dataset developed to study annotator biases *per se* may give little direct insight into online hate. It is important to clearly delineate the abusive phenomena that are and are not addressed by a particular task definition. In the remainder of this section, we provide a framework for explicating and delineating detection tasks. We explore the two most salient dimensions: (1) the nature of abuse and (2) the granularity of the taxonomy.

**Detection tasks: The nature of abuse.** The nature of the abuse refers to *what* is targeted/ attacked in the abusive content. The most well-established taxonomic distinction in this regard is the difference between (i) interpersonal abuse, and (ii) group-directed abuse [22]). Other authors have sought to deductively theorise additional categories, such as 'concept-directed' abuse, although these have not been widely adopted [11]. Research by Salminen et al. shows the possibility of integrating analysis of concepts with abuse detection, such as by analysing the prevalence and dynamics of toxic language across online comments relating to different news topics [54].

Through an inductive investigation of existing training datasets, we extend the widespread binary distinction between person- and group- directed abuse to four primary categories, as well as a fifth 'Mixed' category.

1. *Person-directed abuse.* Content which directs negativity against individuals, typically through aggression, insults, intimidation, hostility and trolling, amongst other tactics. Most research falls under the auspices of 'cyber bullying', 'harassment' and 'trolling' [40, 55, 56]. One of the most widely used datasets contains 'personal attacks' found in English Wikipedia editor comments [47] and another focuses on trolling, distinguishing between 'direct harassment' and other behaviours [57]. Important considerations in studies of person-directed abuse are (a) the interpersonal relations between actors, such as whether individuals engage in patterns of abuse or one-off acts and whether they are known to each other in the 'real' world (both of which are a key concern in studies of cyberbullying) and (b) standpoint, such as whether individuals directly engage in abuse themselves or encourage others to do so. For example, the theoretically sophisticated synthetic dataset provided by [56] identifies not only harassment but also *encouragement* to harassment. Similarly, [39] mark up posts from computer game forums (World of Warcraft and League of Legends) for cyberbullying and annotate these as ⟨offender, victim, message⟩ tuples.

2. *Group-directed abuse.* Content which directs negativity against a social identity, which is defined in relation to a particular attribute (e.g. ethnic, racial, religious group membership) [58]. Such abuse is often directed against marginalised or under-represented groups in society. Group-directed abuse is typically described as 'hate speech' and includes use of dehumanising language, making derogatory, demonising or hostile statements, making threats, and inciting others to engage in violence, amongst other dangerous communications. Common examples of group-directed abuse include sexism, which is included in datasets provided by [32, 56, 59–61] and racism, which is included in [32, 62]. Determining the limits of any group-directed abuse category (e.g. 'Islamophobia') requires careful theoretical reflection about where the boundary of the identity lies given the intersecting nature of most social characteristics. For instance, researchers of racism have to decide whether to include ethnic, caste-based and certain religious prejudices. There is no 'right' answer to such questions as they engage with ontological questions about the nature of identity and the politics of categorization. In some cases, datasets are marked up just for hate speech *per se* rather than against a specific group.

3. *Flagged content.* Content which is reported by community members or assessed by community and professional content moderators to be abusive and/or undesirable. Ljubešić et al. provide two large datasets with flagged content from news providers, Slovenia's RTV MCC and Croatia's 24sata [42]. Both are available, albeit in encrypted form. Datasets which contain flagged content are usually broad as moderators also remove spam, sexually inappropriate content and other undesirable contributions.

4. *Incivil content.* Content which is considered to be incivil, rude, inappropriate, offensive or disrespectful [41, 42, 63]. Such categories are usually defined with reference to the *tone* that the author adopts rather than the *substance* of what they express. Such content usually contains obscene, profane or otherwise 'dirty' words. This type of abuse can be easier to detect as closed-class lists are effective at identifying single objectionable words (e.g. [64]). However, one concern with such approaches is that the presence of 'dirty' words does not necessarily signal malicious intent or abuse; they may equally be used as intensifiers or colloquialisms [65]. Detecting incivility can be particularly difficult if it requires annotators to infer the subjective intent of the speaker or to understand (or guess) the social norms of a setting and thus whether disrespect has been expressed [66].

5. *Mixed*. Content which contains multiple types of abuse, usually a combination of the other four categories. Integrating the study of different types of abuse is important because in the real world incivil, hateful and harassing content often overlaps. For instance, female politicians may receive more interpersonal abuse than other politicians. This might not appear as misogyny because their identity *as women* is not referenced—but their gender might have nonetheless *motivated* the abuse they were subjected to. Mixed forms of abuse require further research, and have thus far been most fully explored in the OLID dataset provided by [4].

**Detection tasks: The granularity of taxonomies.** The granularity of the taxonomy refers to how much detail it contains. The most important and widespread distinction is whether a binary classification is used (e.g. Hate, Not) or a multi-level classification, such as a tripartite split (e.g. Overt, Covert and None). In some cases, a large number of complex classes are created, such as combining what entity is targeted, along with its theme and strength.

In general, social scientific research encourages creating a detailed taxonomy with a large number of fine-grained categories, such as distinguishing between different types of explicit abuse as well as a range of more subtle behaviours, such as micro-aggressions [11]. However, it is an open-question as to whether this level of detail aids in machine learning work. A more complex taxonomy is only suitable for machine learning if there are enough data points in each category and if annotators are capable of consistently distinguishing between them [57, 60, 67]. Complex annotation schemas may not result in better training datasets if they cannot be implemented in a robust and consistent way by annotators. This design issue has practical ramifications; detection systems trained on datasets which do not distinguish between conceptually distinct categories might have low performance because of considerable within-class variation and might not be usable for downstream tasks [27, 68]—but, at the same time, classifiers trained on datasets with insufficient data points might have very low performance and be equally unusable. Dataset creators must balance these potentially competing priorities.

**Detection tasks: A hierarchy of taxonomic detail.** Drawing together our remarks on both (1) the nature of abuse and (2) the granularity of taxonomies, we identify a hierarchy of taxonomic granularity from least to most detailed:

1. Binary classification of a single 'meta' category, such as hate/not or abuse/not. This can lead to very general and vague datasets and classifiers which may be inappropriate in many applications. For instance, Alakrot et al. provide an Arabic dataset which is labelled for 'a range of offensive language and flaming' [69] [p. 174].

2. Binary classification of a single type of abuse, such as person-directed or group-directed. This can be problematic given that abuse is nearly always directed against *a group* rather than *'groups' per se*. For instance, Alfina et al. provide an Indonesian dataset which is labelled for any form of identity-based hate [70] [p. 174].

3. Binary classification of abuse against a single well-defined group, such as racism/not or Islamophobia/not, or interpersonal abuse against a well-defined cohort, such as elected politicians or young people. For instance, Ross et al. provide a German dataset which identifies xenophobic abuse [71].

4. Multi-class classification of different types of abuse, such as:

   a. Different targets of abuse. For instance, Waseem and Hovy provide a dataset with annotations for sexist, racist and neutral content [32].

   b. Different strengths of abuse. For instance, Davidson et al. provide a dataset with annotations for hateful, offensive and neutral content [27].

   c. Different types of statements and themes. For instance, Chung et al. provide a dataset for Islamophobia which also contains annotations for themes, including culture, economy and criminality [72].

5. Multi-class classification of different focuses of abuse (e.g. person and group- directed) which is also integrated with other dimensions of abuse (such as the strength and type of statement). For instance, Ousidhoum et al. provide datasets in English, French and Arabic with annotations for the hostility, directness, the target attributes, the target group and the annotator sentiment [73]. Not that the amount of unique combinations increases exponentially as more annotation dimensions are added, which quickly requires a large amount of data points, potentially making the classification task intractable.

## The content of training datasets

The content of the datasets varies considerably across the 63 we have analysed. Drawing on the data statements framework, relevant social theory and other literatures, we identify several factors which can be used to understand the content of datasets.

**The 'level' of content.** Of the 63 training datasets, 62 are annotated at the level of the post and only one dataset is annotated at the level of the user [74]. None of them are annotated at the level of the comment thread. Only three of the datasets suggest that the entire conversation thread was presented to annotators when marking up each entry, meaning that in most cases this important contextual information is not used. 62 of the training datasets contain only text, with just one containing images [75]. This is a considerable limitation of existing research [11], especially given the multimodal nature of online communication and the increasing ubiquity of digital-specific image-based forms of communication such as Memes, Gifs, Filters and Snaps [76]. Although some work has addressed the task of detecting hateful images [75, 77, 78], this has not led to publicly available datasets. Note that since we conducted this review, one dataset has been subsequently released which contains hateful posts with both text and image, and promoted through a well-funded competition [79]. To our knowledge, no published research has tackled the problem of detecting hateful audio content. This is a distinct challenge; audio contains contains important vocal cues which provide more opportunities to investigate (but also potentially misinterpret) the tone and intention behind abuse.

**Language.** The most common language in the training datasets is English, which appears in 25 datasets, followed by Arabic (6 datasets), Italian (5 datasets), then German and Hindi-English (4 datasets each). The languages of all 63 training datasets is shown in S3 Fig. Noticeably, major languages, both globally and in Europe, do not appear, which suggests considerable unevenness in the linguistic and cultural focuses of abusive language classification. For instance, languages such as Farsi and Russian do not appear at all and there are major gaps in the coverage of European languages, including Finnish and Dutch. Greater diversity of languages and dialects in training datasets would support efforts to develop systems for multilingual abusive content detection. This is a promising area given the availability of high-performing general-purpose NLP models for multilingual content, such as mBert, XLM-R, and m-USE [80], although these are yet to be fully exploited for abusive content. These efforts are further supported by high-profile competitions and events, such as Kaggle's multilingual detection competition, sponsored by Jigsaw with a large cash prize [81]. Better performance in this area would have significant real world consequences; detection systems for violent content in less well-known dialects and languages could be used to efficiently identify (and de-escalate) risks of inter-group conflict globally.

**Source of data.** Training datasets sample data from a range of online spaces, including mainstream platforms, such as Twitter [32] and Wikipedia [47], as well as more niche forums, such as World of Warcraft [39] and Stormfront [82]. In most cases, data is collected from public sources and then manually annotated but in others data is sourced through proprietary data sharing agreements with host platforms [41]. The sources of the 63 datasets is shown in S4 Fig.

Unsurprisingly, Twitter is the most widely used source of data, accounting for 39 of the 63 datasets (36 contain only Twitter data and in a further 3 it is combined with Facebook data). This reflects wider concerns in computational social research that Twitter is over-used, primarily because it has a very accessible API for data collection [83, 84]. Facebook appears in 6 datasets (3 of which are combined with Twitter data), followed by Wikipedia which is used for 3 datasets—although we note that all of these were reported in the same publication. Many of the most widely used online platforms appear in only one dataset, such as Reddit and YouTube [69, 85] or are not represented at all, such as Weibo and VK. There are two datasets for gaming platforms (World of Warcraft and League of Legends [39]) but, overall, gaming sites are not well represented in the training datasets. This is surprising given concerns about gaming communities' links with hateful and toxic behaviours [86]. The lack of diversity in where data is collected from limits the development of generalisable and robust detection systems. We identify three main issues:

1. Linguistic practices vary across platforms. Twitter only allows 280 characters (previously only 140) and abusive content detection systems trained on this data may not perform as effectively when faced with longer pieces of text. More broadly, the technical affordances of platforms can affect the style, tone and topic of the content they host, making cross domain application of classification systems difficult.

2. The demographics of users varies across platforms. Social science research indicates that 'digital divides' exist, whereby online users are not representative of wider populations and may vary between different online spaces [84, 87, 88]. Blank draws attention to how Twitter users are usually younger and wealthier than offline populations [88]. He also highlights important cross-national differences: British Twitters are better-educated than the offline British population but the same is not true for American Twitter users compared with the offline American population. These demographic differences are likely to affect the types of content that users produce and how they express themselves online.

3. Platforms have different norms and so host different types and amounts of abuse. Mainstream platforms have made efforts in recent times to 'clean up' content and so the most overt and aggressive forms of abuse, such as direct threats, are likely to be taken down [89]. However, more niche platforms, such as Gab or 4chan, tolerate more offensive forms of speech and are more likely to contain explicit abuse, such as racism and very intrusive forms of harassment, such as 'doxxing' [6, 90, 91].

**Size of datasets.** The size of the training datasets varies considerably from 469 posts to 17 million; a difference of five orders of magnitude. Differences in size partly reflect different designs and annotation practices, and the largest datasets are usually from proprietary data sharing agreements with platforms. Smaller datasets tend to be carefully collected and then manually annotated. The size of the 63 datasets is shown in S5 Fig. Smaller datasets are problematic because they contain too little linguistic variation and increase the likelihood of overfitting. It can also lead to worse performance: Rizoiu et al. [92] train detection models on only a proportion of the Davidson et al. and Waseem training datasets and show that this leads to lower performance, particularly for 'data hungry' deep learning approaches [92]. However, there are no established guidelines for how large an abusive training dataset needs to be—and 'big' datasets alone are not a panacea for the challenges of abusive content classification. Large training datasets which have been poorly sampled, use theoretically problematic categories or are inexpertly and unthoughtfully annotated, could still lead to poor classification systems.

The challenges posed by small datasets could potentially be overcome through machine learning techniques such as 'semi-supervised' and 'active' learning [93], although these have only been limitedly applied to abusive content detection so far [94]. Sharifirad et al. propose using text augmentation and new text generation as a way of overcoming small datasets, which is also a promising avenue for future research [95].

**Class distribution and sampling.** Class distribution is an important, although often under-considered, aspect of the design of training datasets. Datasets with little abusive content will lack linguistic variation in terms of what is abusive, thereby increasing the risk of overfitting. More concerningly, the class distribution directly affects the nature of the engineering task and how performance should be evaluated. For instance, if a dataset is 70% hate speech then a zero-rule classification system (i.e. where everything is categorised as hate) will achieve 70% precision and 100% recall. This needs to be accounted for when evaluating performance: all else being equal, 80% precision on an evenly balanced dataset is more impressive than a dataset with 70% hate. Considering the class distribution is particularly important when evaluating the performance of ternary classifiers which tend to have greater imbalances.

On average, 36.7% of the content in the training datasets is abusive. Note that each dataset is weighted equally. In cases where datasets contain multiple categories, we assess the class distribution based on any tweets which are not non-abusive, for instance combining tweets which are abusive, profane and hateful. Where different distributions are reported because of different levels of agreement, we take the distribution based on 2/3 agreement, as this is the most widely used across the field. Class distributions vary considerably, from those with just 1% abusive content up to 100%. Of the four datasets that comprise 100% abusive content, three are synthetic datasets (reported in one publication [72]) and the other is an amendment to an existing dataset and only contains sexism [59]. The percentage of content which is abusive in the 63 datasets is shown in S6 Fig.

The class distribution is largely a product of how data was sampled and which platform. Bretschneider [39] created two datasets without using purposive sampling, and as such they contain very low levels of abuse (1%). Other studies filter data collection to increase the prevalence of abuse, sampling purposively using content from a certain time period, community, keywords/hashtags or individuals. Purposive sampling has been criticised for introducing various forms of bias into datasets [96], such as missing mis-spelled content and only focusing on the linguistic patterns of an atypical subset of users [97].

Sampling strategy is a question of both class distribution and dataset integrity. Davidson et al. argue that racial biases are partly due to oversampling of certain keywords [19]—given that most researchers do not know the true prevalence of the keywords they use to sample content, datasets can become skewed and biased in ways that the creators are not fully aware of. Equally, using a limited time window for sampling (especially one which may relate to a specific event) can severely limit the generalisability of any systems trained the dataset. A related risk is that at present more data is sampled from far right than far-left communities [82, 98]—which means that most hate speech classifiers implicitly pick up on right wing styles of discourse rather than just the core features of hate speech. This could have profound consequences for our understanding of online political dialogue if the classifiers are applied uncritically to other groups, such as left-wing communities.

The prevalence of abuse 'in the wild' is very low, likely less than 1% depending on the community [99]. This means that nearly all of the 63 training datasets are radically unrepresentative and that how they are evaluated (i.e. on datasets with a large number of positive cases) is not comparable to the settings in which they are likely to be used. This shows the importance of deciding the *purpose* of the dataset before its creation. For instance, a dataset that is suitable for maximizing classification performance on an out-of-sample use case (e.g. in a real content moderation setting) may not be suitable for understanding or explaining classification results. In the first case a more representative class distribution might be more appropriate but in the second an equal split could be better. Arguably, more thought needs to be given to what class distribution is desirable for a given task, rather than just thinking that 'more abuse is better'.

**Identity of the content creators.** The identity of the users who originally created the content in training datasets is fully described in only four cases (reported on in two papers) [56, 72]. In all cases the data is synthetic. Chung et al. use 'nichesourcing', getting experts in tackling hate speech to create hateful posts [72]. Sprugnoli et al. ask children to adopt pre-defined roles in an experimental classroom setup, and to engage in a cyberbullying scenario [56]. In most of the datasets, the identity of the content creators, such as their identity, demographics, online behavioural patterns and affiliations, are not provided.

Providing more information about content creators may help address biases in existing datasets. For instance, Wiegand et al. show that 70% of the sexist tweets in the highly cited Waseem and Hovy dataset [32] come from two content creators and that 99% of the racist tweets come from just one [96]. This is a serious constraint as it means that user-level metadata is artificially highly predictive of abuse. And, even when user-level metadata is not explicitly modelled, detection systems only need to pick up on the linguistic patterns of a few authors to nominally detect abuse. Providing more information about the identity of content creators when datasets are created would help to mitigate this issue. The lack of information may also be driving as-yet-unrecognised biases.

## Annotation of training datasets

**Annotation process.** How training datasets are annotated is one of the most important aspects of their creation. We identify five different annotation processes in the 63 datasets:

1. *Experts/trained workers* (28 datasets, 44%). Annotation by experts or trained workers is the most widely used practice we identify [82, 100, 101]. It is time-intensive but usually produces higher quality annotations. Waseem reports that 'systems trained on expert annotations outperform systems trained on amateur annotations.' [102] [p. 138] and, similarly, D'Orazio et al. claim, 'although expert coding is costly, it produces quality data.' [103]. However, the notion of an 'expert' and what counts as 'training' is fuzzy within abuse research. In many cases, publications only report that 'an expert' or trained coder is used, without specifying the nature of their expertise or level of training—even though this can vary substantially.

For example, an 'expert' may refer to an NLP practitioner, an undergraduate student with only modest levels of training, a member of an attacked social group relevant to the dataset or a researcher with a doctorate in the study of prejudice. In general, we anticipate that experts in the social scientific study of prejudice/abuse would perform better at annotation tasks then NLP experts who may not have any direct expertise in the conceptual and theoretical issues of abusive content annotation. In particular, one risk of using NLP practitioners, whether students or professionals, is that they might 'game' training datasets based on what they anticipate is technically feasible for existing detection systems. For instance, if existing systems perform poorly when presented with long range dependencies, humour or subtle forms of hate (which are nonetheless usually discernible to human readers) then NLP experts could unintentionally use this expectation to inform their annotations and not label such content as abusive.

2. *Crowdsourcing* (21 datasets, 33%). Crowdsourcing is the second most widely used approach [73, 85, 104]. It is widespread in NLP research because it is relatively cheap and easy to implement. The value of crowdsourcing lies in having annotations undertaken by 'a large number of non-experts' ([105], p. 278)—any bit of content can be annotated by many (most likely unskilled) annotators, effectively trading quality for quantity. However, studies which use crowdsourcing with only 3 annotators for each bit of content risk minimising quality without counterbalancing it with greater quantity. Furthermore, testing and evaluating the work of many different annotators can be challenging [106, 107] and ensuring they are paid an ethical amount may make the cost comparable to using trained experts. Crowdsourcing has also been associated with 'citizen science' initiatives to make academic research more accessible but this may not be fully realised in cases where annotation tasks are laborious and low-skilled [50, 108].

3. Professional moderators (3 datasets, 5%). Professional moderators offer a standardized approach to content annotation [41, 42]. This should, in principle, result in high quality annotations. However, one concern is that moderators are too output-focused as their work involves determining whether content should be allowed or removed from platforms; they may not provide detailed labels about the nature of abuse and may also set the bar for content labelled 'abusive' fairly high, missing out on more nuanced and subtle varieties. In most cases, moderators will annotate for a range of unacceptable content, such as spam and sexual content, and this must be marked in datasets.

4. A mix of crowdsourcing and experts (7 datasets, 11%). This can take a range of forms, including multi-stage annotation designs, mixing different approaches to make annotation more scalable [102, 109, 110].

5. Synthetic data creation (4 datasets, 6%). Synthetic datasets are an interesting and, if created carefully, potentially very effective option. They are inherently non-authentic and therefore not necessarily representative of how abuse manifests in the real-world. However, if they are created in realistic conditions by experts or relevant groups then they can mimic real behaviour and have the added advantage that they may have broader coverage of different types of abuse. Synthetic data can lead to better understanding of model limitations and robustness, such as by 'perturbing' or 'contrasting' abuse to produce subtly different varieties or to adjust the content just enough to change the primary label [111]. Synthetic datasets are also usually easier to share.

**Identity of the annotators.** The data statements framework given by Bender and Friedman emphasises the importance of understanding who has completed annotations because 'their own "social address" influences their experience with language and thus their perception of what they are annotating' [48] [p. 591]. This can include their socio-demographic traits [12] as well as their native language and cultural background. For instance, the dataset provided by Founta et al. used annotators from Venezuela, a non-English speaking majority country, to mark up content in English, potentially introducing biases [67].

Of equal importance as the profile of each annotator is the diversity of the overall pool. A homogeneous group will be poorly equipped to catch all instances of abuse in a corpus because no annotator will be well-versed in *all* of the slang or coded meanings used to express abuse [112]. Indeed, many of these coded meanings are deliberately covert and obfuscated [113].

Information about annotators is unfortunately scarce in the 63 datasets. In 27 no information is given about the identity of annotators [27]; in 24 datasets very limited information is given [73], such as whether the annotator is a native speaker of the language; and in just 12 cases is detailed information given [47]. Interestingly, only 4 out of these 12 datasets are in the English language.

One challenge facing researchers is that precisely what information is needed to evaluate and understand the impact of annotator differences is unclear, and guidelines have not been previously established. As a first step, we split information about annotators into (i) Demographic information, (ii) subject-matter expertise and (iii) personal experiences. That said, we caution that more research is needed to fully understand the role of different annotator perspectives; Waseem and Hovy used self-identified anti-racist activists and feminists to mark up content but their dataset was subsequently found to still contain substantial racial biases [19].

1. Demographic information. The nature of the task will affect what information should be provided, as well as the geographic and cultural context. For instance, research on Islamophobia should include, at the very least, information about annotators' religious affiliation. Relevant variables include:

   a. Age

   b. Ethnicity and race

   c. Religion

   d. Gender

   e. Sexual Orientation

2. Expertise and experience. Relevant variables include:

   a. Field of research

   b. Years of experience

   c. Research status (e.g. research assistant or post-doc)

3. Personal experiences of abuse. In our review, none of the datasets contained systematic information about whether annotators had been personally targeted by abuse or had viewed such abuse online, even though this affects annotators' perceptions. Relevant variables include:

   a. Experiences of being targeted by online abuse

   b. Experiences of viewing online abuse

**Guidelines for annotation.** A key variation across datasets is whether annotators were given detailed guidelines, very minimal guidelines or no guidelines at all. Analysing this issue is made difficult by the fact 27 of the 63 datasets do not provide the guidelines (43%) and 20 only provide them in a highly summarised form (32%). Detailed guidelines are given for just 16 datasets (25%). More information about annotation would not only help future researchers to better understand what datasets contain but also to investigate ways of improving and extending them.

Some dataset creators construct clear and explicit guidelines in an attempt to ensure that annotations are uniform and align closely with social scientific concepts. Davidson et al. provide a clear distinction between 'hate' and 'offence', encouraging annotators to clearly separate use of hateful terms from actual hate [27]. In other cases, dataset creators allow annotators to apply their own perception. For instance, in their Portuguese language dataset, Fortuna et al. ask annotators to 'evaluate if according to your opinion, these tweets contain hate speech' [60]. The risk here is that authors' perceptions may differ considerably; Salminen et al. show that online hate interpretation varies considerably across individuals [114].

Inter-annotator agreement scores for abusive content are often low, particularly for tasks which deploy more than just a binary taxonomy. However, this is not necessarily a 'problem' as much a research opportunity; abusive content annotation is better understood, ontologically, as an intersubjective process in which agreement is constructed, rather than an objective process in which a 'true' annotation is 'found'. Because of the inherently subjective, contextual and political nature of abuse, some researchers have shifted the question of 'how can we achieve the correct annotation?' to 'who should decide what the correct annotation is?' [102].

Some aspects of abusive language present particularly significant problems for annotators, such as Irony, Calumniation and Intent. These are fundamentally indeterminate issues, and although annotators can be left to apply their own judgement, providing guidelines (or at least offering a position on how they should be handled) helps to maximize consistency. Indeed, these three issues are all closely related. For instance, Kenny et al. [115] note how sarcasm, irony, and humour complicate the picture of intent by making it difficult to discerning the *true* intent of speakers.

*Irony*. Ironic statements have a meaning contrary to what one might understand at first reading. Lachenicht [116] notes that Irony goes against Grice's quality maxim, and as such Ironic content requires closer attention from the reader as it is prone to being misinterpreted. Irony is a particularly difficult issue with abusive content as in some cases it is used humorously (and thus ironic content might legitimately be considered non-abusive, as with content that lampoons hateful bigotry) but in other cases it is used as a way of *veiling* genuine abuse. Previous research suggests that irony is widespread amongst abusive content. For instance, Sanguinetti et al. [109] find irony in 11% of hateful tweets in Italian. [42] find that irony is common in self-deleted comments; it appears in 33.9% of deleted comments in a Croatian comment dataset and 18.1% of deleted comments in a Slovene comment dataset. Irony is also one of the most common reasons for content to be re-moderated on appeal, according to Pavlopoulos et al. [41]. Annotating irony (as well as related constructs, such as sarcasm and humour) is inherently difficult. [117] report that agreement on sarcasm amongst annotators working in English is low, something echoed by annotations of Danish content [118].

*Calumniation*. Calumny refers to false statements, slander and libel. Amongst the 63 datasets, this is annotated in datasets for Greek [41] and for Croatian and Slovene [42]. Its prevalence varies considerably across these two datasets and reliable estimations of the prevalence of false statements are not widely available. Calumniation raises several conceptual problems: should content that makes highly negative false claims about a person or group be considered abusive? However, if the information is then found out to be true does it make the content any less abusive? And, given the lack of consensus about most issues in a 'post-truth' age [119], who should decide what is considered *true*? How do we determine whether the content creator *knew* whether something was true when they created it? And does it even matter whether they knew it was false? Understandably, most datasets do not taken any perspective on the truth and falsity of content. This is a practical solution: given error rates in abusive language detection as well as error rates in fact-checking, a system which combined both could be inapplicable in practice. However, the ontological, epistemological and social issues of truth and falsity in relation to abusive content a remarkably omission from the field and require further investigation.

*Intent*. Intent refers to the state of mind and motivation of the person who engaged in abuse. It is also a core part of many social scientific definitions of abusive language [120]. Intent is usually used to emphasize the moral *wrongness* of abusive behaviour by showing that it was produced to harm another individual [55, 57, 109]. However, it is difficult to discern the intent of another speaker in a verbal conversation between humans, and even more difficult to do so through written and computer-mediated communications [121]. Most of the guidelines for the 63 datasets do not contain explicit guidelines for intent, although there are exceptions. [40] define subtypes of misogynistic behaviour as "intent to physically assert power over women". [57] state that messages that are "unapologetically or intentionally offensive" fit in the highest grade of trolling under their schema. [122] include intent in their annotation standard and argue that understanding context is crucial to achieving quality annotations (such as by viewing an abusive speakers' other online messages). However, this proposition poses conceptual challenges given that people's intent can shift over time.

## Dataset sharing

### Challenges and opportunities of achieving open science

Sharing data is an important aspect of open science, and an ethical, secure and accessible mechanism for dataset sharing would provide several scientific benefits to the field. It would (1) enable greater collaboration amongst researchers, (2) enhance the reproducibility of research [123–125] and (3) substantively advance the field by enabling researchers to better understand the limitations of existing research and to identify new research directions. For instance, Binns et al. use the detailed metadata in the datasets provided by Wulczyn et al. to investigate how the demographics of annotators impacts the annotations they make [12, 47]. The value of such insights is only clear after the dataset has been shared—and, equally, is only possible because of data sharing. Furthermore, open sharing of data could enable more commercial innovation, generating additional economic and social benefits.

Open sharing of datasets is not only a question of scientific integrity and advancing scientific knowledge. It is also, fundamentally, a question of fairness and power. Opening access to datasets will enable less-well funded researchers and organisations, which includes researchers in the Global South and those working for not-for-profit organisations, to steer and contribute to research. This is a particularly pressing issue in a field which is directly concerned with the experiences of often-marginalised communities and actors [32].

Dataset sharing poses ethical and legal challenges, especially in light of recent regulatory changes, such as the introduction of GPDR in the UK [126, 127]. This problem is particularly acute with abusive content, which can be deeply shocking, and some training datasets from highly cited publications have not been made publicly available [46, 128, 129]. Open science initiatives can also raise concerns amongst the public, who may not be comfortable with researchers sharing their personal data [130, 131]. At present, most researchers do not obtain explicit consent from the users who content they are analysing and, instead, rely on the implicit consent that users are in a public or semi-public space [132]. Even if the risk of harm is minimized this can raise concerns, especially where users' are not anonymous and are being associated with abusive content. In addition, some researchers may be unsure about what data can be shared under platforms' Terms of Service, and may receive conflicting advice from their host institution. To our knowledge, no successful initiatives for sharing abusive training datasets have emerged in the field, beyond ad-hoc use of platforms such as Github and Zenodo or through sharing data in competitions. This is a widespread problem in computational research [133].

There are two key challenges which must be overcome to ensure that training datasets can be shared and used by future researchers. First, dataset integrity must be maintained. At present, several of the most widely used datasets provide only the annotations and IDs and must be 'rehydrated' to collect the content. The dataset provided by Waseem and Hovy must be collected in this way [32], and has degraded considerably since it was first released as the tweets are no longer available on Twitter. Chung et al. also estimate that within 12 months the dataset for counterspeech by Mathew et al. had lost more than 60% of its content [6, 72]. Dataset degradation poses several risks: First, if less data is available then there is a greater likelihood of overfitting. Second, the class distributions usually change as proportionally more of the abusive content is taken down than the non-abusive. Third, it is also likely that the more overt forms of abuse are taken down, rather than the covert instances, thereby changing the qualitative nature of the dataset.

Second, dataset access must be securely permissioned so that researchers can use them whilst respecting platforms' Terms of Service and barring access to malicious actors. The difficulty of sharing data in sensitive areas of research is reflected in the history of the website 'Jihadology'. Its original purpose was to provide access to research materials for studying Islamist extremism, such as ISIS propaganda. In 2019, counter-terrorism officials from the UK's Home Office attempted to shut it down, after which it restricted public access. Officials were concerned that, whilst it aimed to support academic research into Islamist extremism, Jihadology may have inadvertently enabled individuals to radicalise by making otherwise banned extremist material available. By working with partners such as the not-for-profit Tech Against Terrorism, Jihadology created a secure area in the website, which can only be accessed by approved researchers. However, this is an expensive and custom-build solution that may not be available to all dataset creators.

Several solutions are available to tackle these problems.

1. *Synthetic datasets*. From a legal perspective, synthetic datasets can be shared freely because they do not face any restrictions from platforms' Terms of Service. However, (a) this can introduce other biases and may not be a sustainable solution and (b) there are still other ethical- and social- challenges in sharing such content. Synthetic datasets still need to be shared in a way to limit access for potential extremists but face no challenges from Platforms' Terms of Services. Ultimately, creating synthetic data is not a scalable solution for the whole field to adopt; only four of the 63 datasets we have reviewed were developed synthetically.

2. *Data 'philanthropy' or 'donations'*. Data philanthropy is defined as 'the act of an individual actively consenting to donate their personal data for research' [131]. This would minimize some ethical and legal risks in abusive dataset sharing as individuals would provide explicit consent. However, this poses several challenges. Many individuals who share abusive content may be unwilling to 'donate' their data, creating severe class imbalances and other biases [131]. Data donations could also open new moral and ethical issues; individuals' privacy could be impacted if their data is re-analysed to derive new unexpected insights [134]. Informed consent is difficult given that the exact nature of analyses may not be known in advance. Finally, data donations alone do not solve how access can be responsibly protected and how platforms' Terms of Service can be met.

3. *Platform-backed datasets.* Platforms could share datasets and support researchers' access. To our knowledge, there are no working examples of this approach in abusive content detection research, but it has been successfully used in other research areas. For instance, Twitter has made available a large dataset of accounts linked to malicious information operations, known as the "IRA" dataset (Internet Research Agency). This would require considerably more interfaces between academia and industry, which may be difficult given the challenges associated with existing initiatives, such as Facebook's Social Science One. However, in the long term, we propose that this is the most effective solution for sharing training datasets. It would remove Terms of Service limitations and it would give researchers access to the large volumes of original content they hold, much of which has moderation metadata. Platforms also have the financial resources to easily fund such an endeavour. We propose two ways of platform-backed sharing: (1) platforms could make content which has violated their Community Guidelines available directly through a secure platform/API or (2) they could waive the Terms of Service for datasets which researchers have collected publicly—thereby making sure that datasets do not degrade over time. This second option is less preferable as it only resolves some of the issues in dataset sharing.

4. *Data trusts.* Data trusts have been described as a way of sharing data 'in a fair, safe and equitable way' ([135] p. 46). However, there is considerable disagreement as to what they entail and how they would operate in practice [136]. The Open Data Institute identifies that data trusts aim to make data open and accessible by providing a framework for storing and accessing data, terms and mechanisms for resolving disputes and, in some cases, contracts to enforce them. For abusive content training datasets, this would provide a way of enabling datasets to be shared, although it would require considerable institutional, legal and financial commitments.

A data trust is the most promising way of enabling datasets to be shared, especially as this provides a meta-framework in which the other proposed solutions (synthetic data, data philanthropy and platform-backed sharing) could be implemented. Initially, this could take the form of a database, which would contain all of the publicly available abuse training datasets. A data trust would substantially reduce the burden on researchers when finding and accessing data. Once they have been approved to the trust, they could access all of the datasets in go. This would level the academic playing field by making all data available. The data trust would also contain all of the metadata reported with datasets and. A simple API could be developed for reading data, similar to that of the HateBase [137]. Eventually this could be extended to include deposits of new training datasets, at which point all information could be included, based on the 'data statements' work of Bender and Friedman [48].

The data trust would need to be permissioned and access controlled to address concerns relating to privacy and ethics. The permissioning system could be maintained either through a single well-respected institution or, to avoid power concentrating amongst a small group of researchers, through a decentralised blockchain. A further benefit is that, in the future, different levels of permission could be implemented for different datasets, depending on commercial or research sensitivity.

### A new repository of training datasets: Hatespeechdata.com

The resources and infrastructure needed to create a dedicated data trust and API for sharing abuse training datasets are substantial and require buy-in from researchers across the field. In the interim, to encourage greater sharing of datasets, we have launched a dedicated website which contains all of the datasets analysed here: https://hatespeechdata.com. Based on the analysis in the previous sections, we have also provided partial data statements [48]. In addition, the website contains previously published abusive keyword dictionaries, which are not analysed here but some researchers may find useful. Note that the website only contains information/data which the original dataset creators have already made publicly available. It will be updated with new datasets in the future.

## Best practices for training dataset creation

Much can be learned from existing efforts to create training datasets. We identify best practices which emerge at four distinct points in the process of creating a training dataset: (1) task formation, (2) dataset creation, (3) annotation, and (4) documentation.

### Task formation: Defining the task addressed by the dataset

Dataset creation should be 'problem driven' [138], with a clear motivation that addresses a well-defined and specific task. This will directly inform the taxonomy design, which should be developed by engaging with relevant social scientific theory. Defining a clear task which the dataset addresses is especially important given the complexity of online abuse, the maturation of the field, and ongoing terminological disagreement. The diversity of phenomena that fits under the umbrella of abusive language means that 'general purpose' datasets are unlikely to advance the field as a whole. Clarity, concision and precision are crucial for communicating the task that is being addressed.

### Creating datasets for abusive language annotation

Once the task is established, dataset creators should select what language(s) will be annotated, where data will be sampled from and how sampling will be completed. At present, the field relies too heavily on data from Twitter. Less well-used sources should always be considered first, such as Instagram and Snapchat. Dataset builders should have a specific target size and class distribution, as well as an idea of the minimum amount of data that is needed. This is where steps 1 and 2 intersect: the data selection should be driven by the problem that is addressed rather than what is easy to collect.

Ensuring there are enough instances of abuse will always be challenging as the prevalence of abuse is so low. However, given that purposive sampling inevitably introduces biases, creators should explore a range of options before determining the best one—and consider using multiple sampling methods at once, such as including data from different times, locations, types of users and platforms. Other options include using measures of linguistic diversity to maximize the variety of text included in datasets. Indeed, a promising avenue for research is to construct datasets which intentionally contain certain features, or to produce 'perturbed' or manipulated versions of real content [20, 139]. This helps to retain the 'real-world' traits of newly collected content whilst also ensuring that content is diverse.

### Annotating abusive language

Annotators should be hired based on their skill sets and experience. However, as noted above, of equal importance is the diversity and range of the overall pool of workers. Efforts should be made to minimize biases which can occur when, for instance, only university students or men are hired. Crowdsourced workers may be an appealing option but whether they are truly appropriate should be carefully considered. We propose that, given the complexities of abusive content and the widespread use of complex taxonomies, crowdsourcing should be viewed as the aberration (in need of careful justification each time it is used) rather than the norm.

Once hired, annotators should be given be given emotional and practical support (as well as appropriate financial compensation). The harmful and potentially triggering effects of annotating online abuse should be recognised at all times. For a set of guidelines to help protect the well-being of annotators, see [11]. This may also address issues of bias as a wider range of workers apply for, and continue with, annotation work.

Annotators work best with clear guidelines, which contain easy-to-understand examples [140]. The best examples are both illustrative, in order to capture the concepts (such as 'threatening language') and provide insight into 'edge cases', which is content that only *just* crosses the line into abuse. Decisions should be made about how to handle intrinsically difficult aspects of abuse, such as irony, calumniation and intent (see above). Annotation guidelines should be developed iteratively by dataset creators; by working through the data, rules can be established for difficult or counter-intuitive coding decisions, and a set of shared practices developed. Annotators should be included in this iterative process. Discussions with annotators about the language that they have seen "in the field" will lead to more consistent data and provide a knowledge base to draw on for future work.

## Documenting methods, data, and annotation

The best training datasets are well-documented and clearly explained. Providing as much information as possible can open new and unanticipated analyses and gives more agency to future researchers. For instance, if all annotators' codings are provided (rather than just the 'final' decision), a classifier could be developed in which the recall of *all* annotations is used to decide final labels rather than just the majority decision [112].

Most of the 63 datasets we surveyed have limited methodological descriptions and few provide enough information to construct an adequate data statement. It is crucial that dataset creators are forthcoming about their biases and limitations: every dataset is biased and this is only problematic when the biases are unknown or hidden. This is particularly important given the intersubjective nature of online abuse and the important role played by political and contextual factors. Without knowing who has given a set of annotations, it can be hard to situate the results, and to identify the appropriate uses (and limitations) of any system trained on them. This is a substantial concern with crowdsourced content where many providers have non-representative workforces, who may lack cultural or domain-specific knowledge, and only minimal metadata about them is provided. Ultimately, given concerns that many societies are now highly polarised, it may be more useful to build systems that detect abuse for a particular context or setting, rather than generic classifiers that detect 'average' abuse.

Full details about the dataset creation process should be documented and provided. For instance, if the task is crowdsourced, then a screenshot of the micro-task presented to workers should be included, and the top-level parameters should be described (e.g. number of workers, maximum number of tasks per worker, number of annotations per piece of text) [50]. Similarly, interface design can be highly influential [141]—if a dedicated interface is used for annotation, this should also be described in prose and shown with a screenshot.

## Implications for practitioners

The typical route for creating a new machine learning classifier is to find data, train a classifier using that data, and to deploy the classifier. There are concrete steps practitioners can take at each of these three stages to improve the quality of a deployed abusive language classifier, and avoid a simplistic 'plug and chug' approach.

A suitable dataset must be selected. When choosing a dataset to use for abusive language classification, there are four key questions to ask. (1) Task fit: Does the dataset capture and present examples of the target abusive language phenomena? Is it a reasonable resource for the task in hand? Data on racism will be overkill for a profanity filter; data on aggression or irony may not work well for detecting sexism. If this is unclear from the dataset documentation, it's probably not the right data. 2) How closely does the genre/domain of the text match? Consider the length and style of the sentences. Training on long-form commentary is likely to give a model that underperforms on short social media text. (3) Is the class balance as expected? Classifiers often predict a similar class distribution prior to the dataset they are trained on. Training a model on 50% abusive/50% non-abusive data risks increasing the chance that the abusive label will be predicted about half the time. This might not be desired, so pick a suitable class balance. (4) How broad and varied is the data? Training data that doesn't capture many different phenomena, different forms of expression, and different types/demographics of author are unlikely to generalise well, leading to a high false negative rate.

Because abusive language is a broad and varied category, care must be taken during training to get good results. Concrete steps can be taken to achieve this. The increased variety in abusive language means that a validation set is of extra value, because the chance of encountering things not observed in any given training set is higher. To make this match a particular application scenario, consider taking an amount of live, *in situ* data and putting that into the validation (and even training) sets. Abusive language is often linguistically complex and can consist of many parts, so use feature representations that capture context comprehensively. Finally, word lists and word list-based approaches should be avoided: they introduce both false positives and false negatives at the same time.

Post-deployment risk can also be mitigated. Consider the cost of false negatives and false positives in the expected abusive language detection system design, and evaluate and select models accordingly. For example, banning users accidentally may be more damaging to a community's health than letting some abuse slip through; on the other hand, a system that is weak at detecting e.g. racist hate-speech is likely to harm both communities and their members as well as organisational image. It is also important to use a representation that fits the deployment scenario. For example, although multilingual BERT supports dozens of languages, using this for building a language-specific classifier can be sub-optimal [142].

## Best practice summary

Our best practice summary reflects the difficulties of creating high quality training datasets for abusive content. Irrespective of the particular form of abuse that is addressed, better practices in dataset creation could lead to higher quality datasets and more interoperability. More standardization is an important aspiration as research continues to mature, although it must be balanced with enabling research innovation and freedom. We summarise our recommendations in the following points:

1. Bear in mind the purpose of the dataset; design the dataset to help address questions and problems situated in abusive language research. Given that abusive content classifiers are increasingly widely used in the 'real world', researchers should consider how any technology they develop will be used, and the social implications for issues such as privacy, protecting from harm and freedom of speech.

2. Avoid using 'easy to access' data, and instead explore new sources which may have greater diversity. Consider what biases may be created by your sampling method. Twitter is overused in studying online abuse and there are opportunities to research less well-studied platforms, such as Snapchat and Instagram. Determine size based on data sparsity and having enough positive classes rather than 'what is possible'.

3. Establish a clear taxonomy for the task, with meaningful and theoretically sound categories. Conflating identity-based hate and interpersonal abuse is likely to lead to systems which cannot be used in the real-world. Poorly constructed datasets risk inducing models which are not robust to simple manipulations, such as negations or counter speech. Explicit inclusion of more complex linguistic features, such as irony and sarcasm, as well as adversarial content, will help to address these issues.

4. Develop guidelines iteratively with your annotators and then publish them with your dataset. Because abuse is intersubjective, establish where lines fall between abuse categories and either maximize consistency or systematically interrogate biases, such as how annotators' backgrounds and experiences influence their annotations. Given the complexities of abusive content, we recommend using trained and contextually-aware annotators rather than crowdsourcing.

5. Report on every step of the research through a Data Statement. Abusive language is complex and how it is interpreted depends on the background of the researcher and annotators; these must be noted. Document aspects of the process which may not be immediately relevant to your task but which could plausibly assist future researchers. Be explicit about challenges, biases and limitations. Proper documentation will encourage standardization.

## Conclusion

This paper examines 63 publicly available datasets for training abusive content detection systems, providing critical insight into what the datasets contain (and omit), how they have been annotated, and how tasks have been formulated. Based on an evidence-driven review, we provided an extended analysis of ways to make training datasets more readily available and useful, including the challenges and opportunities of open science as well as the need for more research infrastructure. We have also reported on the development of hatespeechdata.com, a new repository for online abusive content training datasets. Finally, we outlined best practices for creation of training datasets for detection of online abuse. We have also specified some practical points that practitioners should reflect on when using datasets to create detection systems. These activities meet the four research aims outlined at the start of the paper. Training detection systems for online abuse is a substantial challenge with substantial social consequences. If the systems developed are to be usable, scalable and operate with as few biases as possible, they need to be trained on the right data: garbage in will only lead to garbage out.

## Supporting information

**S1 Checklist. PRISMA 2009 checklist.**
(DOC)

**S1 Fig. Flow diagram of survey selection processes under PRISMA.**
(PDF)

**S2 Fig. Year in which training datasets were originally published.**
(PDF)

**S3 Fig. Primary language of dataset.**
(PDF)

**S4 Fig. Platform from which data is gathered.**
(PDF)

**S5 Fig. Distribution of dataset sizes.**
(PDF)

**S6 Fig. Relative size of "abusive" data class.**
(PDF)

**S1 Appendix. Survey search sources and keywords.** [143].
(PDF)

**S2 Appendix. Datasets included in the survey.**
(PDF)

## Acknowledgments

The authors thank Mr. Alex Harris for providing research assistance and feedback.

## Author Contributions

**Conceptualization:** Bertie Vidgen, Leon Derczynski.

**Data curation:** Bertie Vidgen, Leon Derczynski.

**Formal analysis:** Bertie Vidgen, Leon Derczynski.

**Investigation:** Bertie Vidgen, Leon Derczynski.

**Methodology:** Bertie Vidgen, Leon Derczynski.

**Project administration:** Bertie Vidgen, Leon Derczynski.

**Writing – original draft:** Bertie Vidgen, Leon Derczynski.

**Writing – review & editing:** Bertie Vidgen, Leon Derczynski.